\documentclass[10pt,twocolumn,letterpaper]{article}

\usepackage{iccv}
\usepackage{times}
\usepackage{epsfig}
\usepackage{graphicx}
\usepackage{amsmath}
\usepackage{amssymb}
\usepackage{subfigure}

\usepackage[breaklinks=true,bookmarks=false,colorlinks,linkcolor=red]{hyperref}

\iccvfinalcopy 

\def\iccvPaperID{****} 

\ificcvfinal\fi

\begin{document}

\title{Enriching Local and Global Contexts for Temporal Action Localization}

\author{Zixin Zhu$^1$ ~Wei Tang$^{2}$ ~Le Wang$^{1}$\footnote[1]{}    ~~Nanning Zheng$^1$ ~Gang Hua$^3$ \\

$^{1}$Institute of Artificial Intelligence and Robotics, Xi'an Jiaotong University\\
$^{2}$University of Illinois at Chicago\\
$^{3}$Wormpex AI Research\\
}

\maketitle
{\renewcommand{\thefootnote}{*}
\footnotetext{Corresponding author.}}
\ificcvfinal\thispagestyle{empty}\fi

\begin{abstract}
    Effectively tackling the problem of temporal action localization (TAL) necessitates a visual representation that jointly pursues two confounding goals, i.e., fine-grained discrimination for temporal localization and sufficient visual invariance for action classification. We address this challenge by enriching both the local and global contexts in the popular two-stage temporal localization framework, where action proposals are first generated followed by action classification and temporal boundary regression. Our proposed model, dubbed ContextLoc, can be divided into three sub-networks: L-Net, G-Net and P-Net. L-Net enriches the local context via fine-grained modeling of snippet-level features, which is formulated as a query-and-retrieval process. G-Net enriches the global context via higher-level modeling of the video-level representation. In addition, we introduce a novel context adaptation module to adapt the global context to different proposals.  P-Net further models the context-aware inter-proposal relations. We explore two existing models to be the P-Net in our experiments. The efficacy of our proposed method is validated by experimental results on the THUMOS14 (54.3\% at tIoU@0.5) and ActivityNet v1.3 (56.01\% at tIoU@0.5) datasets, which outperforms recent states of the art. Code is available
    at \url{https://github.com/buxiangzhiren/ContextLoc}.
\end{abstract}

\section{Introduction}

\begin{figure}[t]
\centering
  \subfigure[Snippet-level local context]{\includegraphics[width=1\linewidth]{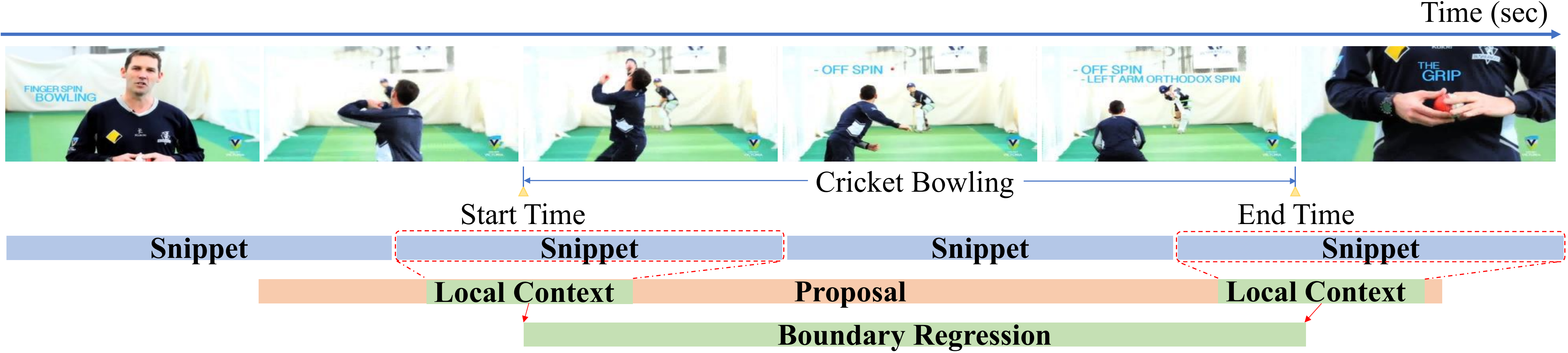}
  \label{fig8}}
  \subfigure[Video-level global context]{\includegraphics[width=1\linewidth]{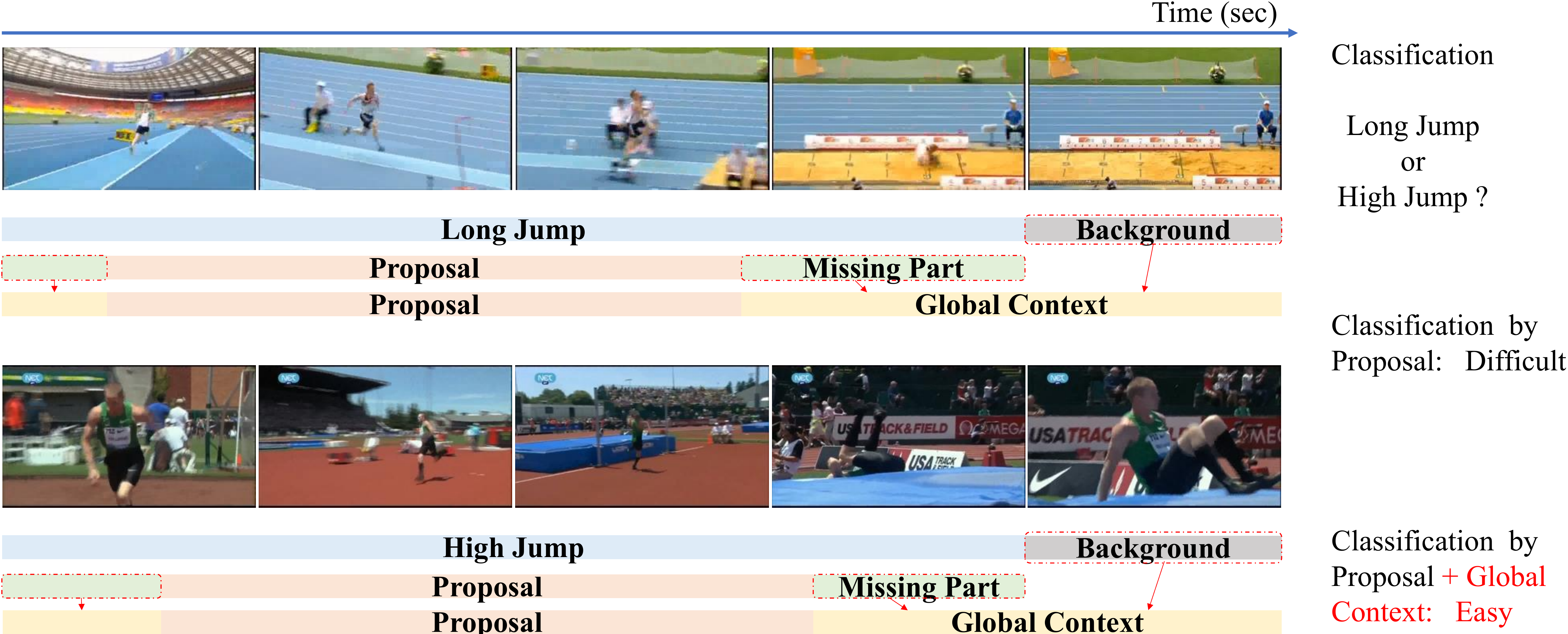}
  \label{fig7}}
\caption{(a) 
Each proposal consists of a number of video snippets. A snippet is a small number of consecutive frames and serves as the basic unit of feature extraction. It is the snippets capturing the start and end times of the action that play an important role in action localization. (b) The video-level global context is important as it involves background and high-level activity information that can be critical  to  distinguish  action  categories  of  similar  appearance and motion patterns.}
\end{figure}

Temporal action localization (TAL) is a fundamental task in video understanding. 
It aims at classifying action instances in an untrimmed video and locating their temporal boundaries. Recently, TAL has drawn increasing attention from the research community because of its wide range of applications such as action retrieval, video summarization, and intelligent security ~\cite{qiu2019learning,otherareas2,otherareas3,otherareas4}.


Prior TAL methods can be divided into two categories. One-stage approaches ~\cite{Huang2019DecouplingLA,yeung2016end,Buch2017EndtoEndST,long2019gaussian} classify and locate action instances from an input video in a single shot. Two-stage approaches first generate category-agnostic action proposals ~\cite{bmn,Alpher02,onesstage1,onestage2,Gao2017TURNTT,zhao2020bottom,bai2020boundary} and then perform action classification and temporal boundary refinement \cite{Alpher03,twostage1,Xu2017RC3DRC,Chao2018RethinkingTF,Dai2017TemporalCN}\label{smtal} for each proposal. They have their respective advantages. One-stage approaches can be easily trained in an end-to-end fashion while two-stage approaches usually obtain superior performance.  

Effectively tackling the task of TAL necessitates a visual representation that jointly pursues two confounding goals, \ie, fine-grained discrimination for temporal localization and sufficient visual invariance for action classification. This paper addresses this challenge by leveraging rich \textit{local} and \textit{global contexts} in a video, in our proposed two-stage method.

The \textbf{local context} refers to snippets within a proposal. 
They contain fine-grained temporal information that is critical to localization. As shown in Figure \ref{fig8}, we locate the boundary of the action ``cricket bowling" by the moments of bowling and catching the cricket. Therefore, it is the snippets capturing these special moments that facilitate localization in the temporal domain.  However, prior approaches  obtain features of a proposal by applying temporal max pooling to features of snippets within it, which unavoidably discards the fine-grained temporal information.



The \textbf{global context} refers to the entire video. It provides discriminative information complementary to features of a proposal for action classification. 
As shown in Figure \ref{fig7}, to distinguish ``long jump'' and ``high jump''\label{lh jummp}, we need to check not only the last a few frames of the action duration but also background frames outside the duration. 
In addition, the global context provides  high-level activity information enforcing strong prior on the categories of actions that should appear in it. For example, it is unlikely to see sports actions in a video of household activity. Unfortunately, the video-level global context has been largely ignored by existing TAL models.

We introduce a novel network architecture, termed \emph{ContextLoc}, to model local and global contexts in a unified framework for TAL. It consists of three sub-networks: L-Net, G-Net and P-Net.
Inspired by the self-attention \cite{Vaswani2017AttentionIA}, 
L-Net performs a \textit{query-and-retrieval} procedure. But different with the self-attention, the queries, keys and values in our L-Net correspond to different semantic entities, and they are specially designed to enrich the local context. Specifically, the feature vector of a proposal is taken as a \textit{query} to match the \textit{key} feature vectors of snippets within this proposal so that the relevant fine-grained \textit{values} in the local context can be retrieved and aggregated to this proposal.

G-Net models the global context by integrating the video-level representation and features of each proposal. However, a naive concatenation of these two would be insufficient because the former contains not only relevant cues but also irrelevant noise. In addition, the portions of context needed to enhance different proposals are different. To effectively integrate video-level information with proposal-level features, we propose \textit{global context adaptation}. It attends the video-level representation to the local context within each proposal so that the global context is adapted to them respectively.

P-Net models context-aware inter-proposal relations. 
This includes interactions between proposal-level features enhanced by the local context and interactions between global contexts adapted to different proposals.
We use an existing model as P-Net and investigate two candidates: P-GCN~\cite{Alpher03} and the non-local network~\cite{nonlocalnn}. 

It is worth noting our ContextLoc is different from P-GCN~\cite{Alpher03}.
P-GCN only considers inter-proposal relations, and the features of a proposal are obtained via applying temporal max pooling to the features of snippets within it. In contrast, ContextLoc enriches the local context via fine-grained modeling of snippet-level features and enriches the global context via higher-level modeling of the video-level representation. 
We consider P-GCN or its relative as a useful component of our framework, \ie, P-Net, and show that the local and global contexts we advocate are complementary to the inter-proposal relations.

We evaluate our proposed method on two popular benchmarks for TAL. On THUMOS14~\cite{t14}, it achieves $54.3\%$ mAP at tIoU 0.5, which outperforms the previous best method PBRNet~\cite{Liu2020ProgressiveBR}. On ActivityNet v1.3~\cite{Heilbron2015ActivityNetAL}, it achieves $56.01\%$ mAP at tIoU 0.5, which outperforms the state-of-the-art method PBRNet.

The contributions of this paper are summarized below.
\begin{itemize}
    \item To our knowledge, this is the first work attempting to exploit the snippet-level local context and the video-level global context to enhance proposal-level features within a two-stage TAL framework. 
    \item We introduce a novel network architecture, termed \textit{ContextLoc}, consisting of three sub-networks, \ie, L-Net, G-Net, and P-Net. 
    L-Net is the first of its kind to use a proposal to query the snippets within it and retrieve the local context to supplement it with fine-grained temporal information.
    G-Net augments features of each proposal by integrating the video-level representation. We introduce a novel context adaptation process to adapt the global context to different proposals. While P-Net is built on
    existing networks, we show that P-Net, regardless of its instantiation, is complementary to our L-Net and G-Net. Our ContextLoc unifies the respective advantages of these three sub-networks and achieves more effective TAL.  
    \item ContextLoc is superior or comparable to the state-of-the-art performance on two popular TAL benchmarks, \ie, THUMOS14 and ActivityNet v1.3.
\end{itemize}


\begin{figure*}
\begin{center}
\includegraphics[width=1\linewidth]{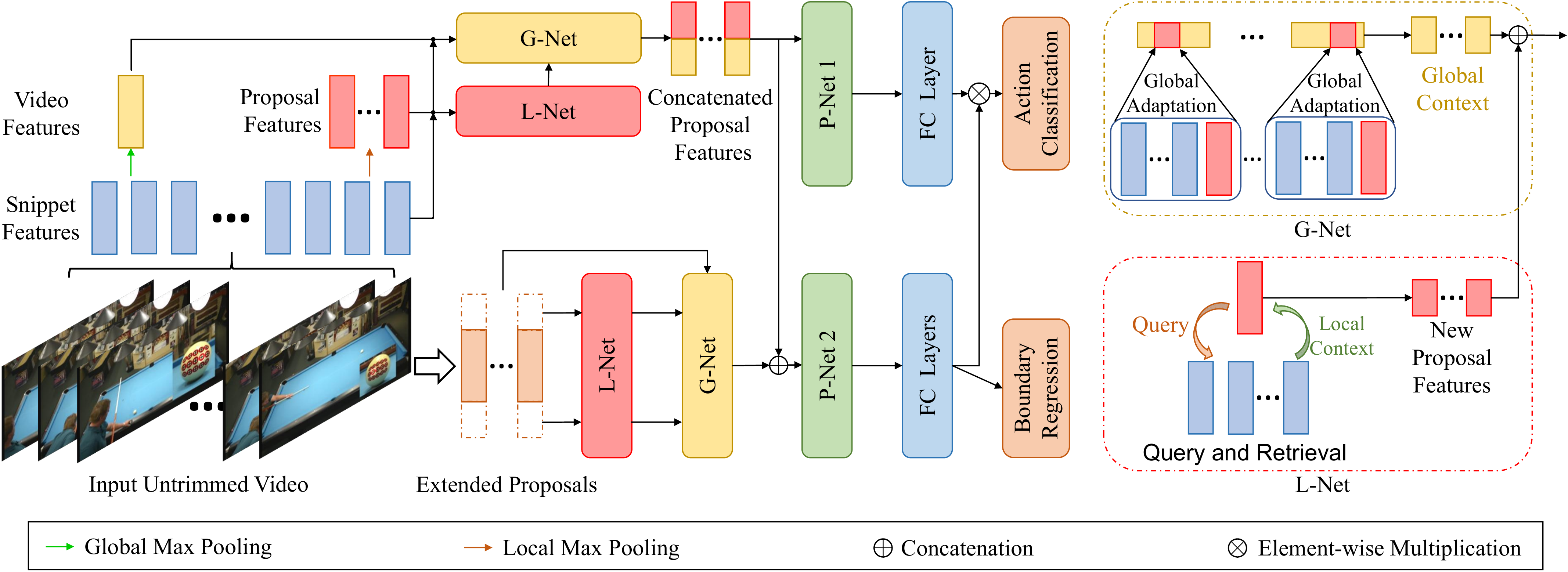}
\end{center}
   \caption{The pipeline of our ContextLoc. 
   The input is an untrimmed video consisting of non-overlapping snippets. Multi-level features are extracted from snippets, proposals, and the video. L-Net, G-Net and P-Net respectively enhance the representation of each proposal via the local context, global context and context-aware inter-proposal relations. Finally, we perform action classification and temporal boundary refinement for each proposal through fully-connected layers. ContextLoc processes both original proposals and extended proposals, and fuses their predictions.}
\label{fig2}
\end{figure*}
\section{Related Work}

\textbf{Action recognition.} Action recognition \cite{Tran2015LearningSF,Tran2018ACL,Wang2013ActionRW,Simonyan2014TwoStreamCN,Gallese1996ActionRI,Ji20133DCN,Alpher01} is a fundamental task in video understanding and has been widely studied in the past a few years. In this paper, we use the I3D network~\cite{Alpher01}, pre-trained on Kinetics~\cite{kay2017kinetics}, to extract snippet-level features from a video.
We use the two-stream strategy like Simonyan \etal~\cite{Simonyan2014TwoStreamCN} to fuse the predictions made on RGB frames and optical flows. There are a few methods exploring the global context for action recognition. He \etal~\cite{he2019stnet} apply temporal convolutions on the local spatial-temporal feature maps to model the global spatial-temporal relationship. Liu \etal~\cite{liu2018global} establish a global temporal template by fusing local motion features over time. 
Different from them, our global context is adapted to each proposal by aggregating the proposal-level and snippet-level features within it for action classification and temporal boundary localization.

\textbf{Temporal action localization.} TAL is a mirror problem of image-based object detection~\cite{chen2020hierarchical,ren2016faster} in the temporal domain. Some methods tackle this task by focusing on video snippets (segments). For example, Zhao \etal~\cite{zhao2017temporal} model the temporal structure of each action instance via a structured temporal pyramid. Shou \etal~\cite{Shou2016TemporalAL} exploit the effectiveness of deep networks in temporal action localization via three segment-based 3D ConvNets. Xu \etal~\cite{gtad} further consider the temporal and semantic structure by  graph convolutional networks. 

Some other works focus on proposals. Lin \etal~\cite{Alpher02} locate temporal boundaries with high probabilities, then directly combine these boundaries into proposals and evaluate whether a proposal contains an action. Chen \etal~\cite{Chen2020RefinementOB} select more reliable proposals and subsequently refine proposal boundaries by variance voting during post-processing. 
Different from these prior TAL methods, we refine proposal-level features with the snippet-level local context and the video-level global context, which they ignore.

Our ContextLoc is different from a recent graph-based method: G-TAD~\cite{gtad}. G-TAD applies a graph convolutional network (GCN) to update features of snippets, each of which is considered as a node in a snippet graph, and casts TAL as a sub-graph localization problem. 
By contrast, ContextLoc enhances features of each proposal by enriching the local snippet-level context and the global video-level context, which are then used for action classification and temporal boundary refinement.

\textbf{Contexts in videos.} Temporal contextual information is important to video understanding and has been exploited in various tasks such as visual relation detection~\cite{Liu_2020_CVPR}, question answering~\cite{fan2019heterogeneous}, and representation learning~\cite{qiu2019learning}. Specifically, Liu \etal~\cite{Liu_2020_CVPR} proposes a novel sliding-window scheme to simultaneously predict short-term and long-term relationships. Qiu \etal~\cite{qiu2019learning} updates local and global features by modeling the diffusions between these two representations through convolutional neural networks. Different from previous work, our exploration of local and global contexts is hierarchical and rich. For proposals, snippets are local, and proposals are global. For videos, proposals are local, and videos are global. The former is modeled by L-Net and the latter is modeled by G-Net. Furthermore, the local and global contexts interact with each other through the P-Net in our proposed model.

\section{The ContextLoc Model}\label{our approach}

\subsection{Overview}
\label{sec:overview}
\textbf{Problem statement.} The input to our ContextLoc is an untrimmed video consisting of $N$ non-overlapping snippets. Each snippet is a small number of consecutive frames. We use the I3D network~\cite{Alpher01} to extract features from each snippet and denote the collection of all snippet-level features as $\{\mathbf{x}_j\in\mathbb{R}^{\textit{D}}\}_{j=1}^N$, where $\mathbf{x}_j$ is the features of the $j$th snippet and $D$ is the feature dimension.
Following the two-stream strategy ~\cite{Simonyan2014TwoStreamCN}, predictions made on RGB frames and optical flows are fused.
The output is a set of action instances $\{\psi_i\ |\ \psi_i=(t_{i,s},t_{i,e},c_i)\}$, where $t_{i,s},t_{i,e}$ and $c_i$ are the starting time, ending time and action category of the $i$th instance, respectively. 

\textbf{Multi-level representations.} As shown in Figure \ref{fig2}, ContextLoc explicitly models representations of three different levels: snippet-level representation $\mathbf{x}\in\mathbb{R}^{{\textit{D}}}$ (local context), proposal-level representation $\mathbf{y}\in\mathbb{R}^{{\textit{D}}}$, and video-level representation $\mathbf{z}\in\mathbb{R}^{{\textit{D}}}$ (global context). The proposals are obtained by BSN~\cite{Alpher02}, and each proposal $i$ has a starting time and an ending time. 
The initial features of the $i$th proposal, denoted as $\mathbf{y}_i$, are obtained via temporally max-pooling the snippet-level features within its duration  $\{\mathbf{x}_j\ |\ j\in{S(i)}\}$, where $S(i)$ denotes snippets whose duration is between the starting time and ending time of the $i$th proposal. 
Similarly, the initial video-level features, denoted as $\mathbf{z}$, are obtained by temporally max-pooling all snippet-level features in the video. 

\textbf{Network architecture.} ContextLoc consists of three sub-networks: L-Net, G-Net and P-Net. They respectively enhance the representation of each proposal via local context, global context, and context-aware inter-proposal relations. 
Following previous work \cite{Shou2017CDCCN,Alpher03,Alpher02,Lin2020FastLO}, 
we also apply ContextLoc to each extended proposal with an enlarged temporal duration.
The final representation of the original proposal is used for action classification (via a fully-connected layer) while that of the extended proposal  is used for completeness prediction and temporal boundary refinement. These classification and completeness scores are element-wisely multiplied to make the final categorical prediction. 

Below we introduce L-Net, G-Net and P-Net in Sections \ref{sec:lnet}, \ref{sec:gnet} and \ref{sec:pnet}, respectively. 
There is a major difference between the way we deal with the extended proposals and how previous approaches do. We detail it in Section \ref{sec:extend}.

\subsection{L-Net (Local Context)}
\label{sec:lnet}
The initial features of a proposal obtained through temporal max-pooling are insufficient because the fine-grained temporal information critical to localization is lost. L-Net addresses this problem by finding within a proposal the snippets most relevant to it and aggregating them to retain the informative features. We call these snippets the local context because their temporal range is within a proposal and they are at a lower semantic level.

Inspired by the self-attention \cite{Vaswani2017AttentionIA}, L-Net performs a \emph{query-and-retrieval} procedure. The \emph{queries}, \emph{keys}, and \emph{values} are respectively features of each proposal, features of snippets within each proposal, and transformed features of these snippets.
Specifically, a \textit{query}  proposal is matched to the \emph{key} snippets within this proposal so that the relevant fine-grained \textit{values} in the local context can be retrieved and aggregated to this proposal.
This is achieved by building an attention map between the query and keys and then aggregating values based on the attention weights.

\textbf{Attention map.} The attention weight between a proposal $i$ and a snippet $j\in S(i)$ measures their relatedness and determines how much information will be retrieved from this snippet. It is calculated as 
\begin{gather}
    a_{i,j}^{L}=\frac{\sigma(s(\mathbf{y}_i,\mathbf{x}_j))}{\sum_{k\in{S(i)}} \sigma(s(\mathbf{y}_i,\mathbf{x}_k))},\label{adj}
\end{gather}
where $\sigma$ is the ReLU function, $s$ is the cosine similarity between two vectors. The ReLU function sets negative similarity values to 0. As a result, those irrelevant snippets will be ignored during the retrieval process.  Normalization ensures the attention weights of all snippets sum to one.

\textbf{Local context aggregation.} 
We first calculate \emph{values} by transforming features of each snippet via a fully-connected layer and then linearly combine them via the attention weights. This local context is finally aggregated with transformed features of the proposal to obtain a new representation of this proposal:
\begin{gather}
	\mathbf{y}^{L}_{i} = \sigma(\mathbf{W}_1^L\mathbf{y}_i +  \sum_{j \in S(i)}a_{i,j}^{L}\mathbf{W}_2^L\mathbf{x}_j),\label{5}
\end{gather}
where $\mathbf{y}^{L}_{i}\in\mathbb{R}^{{\textit{D}}/2}$ is a new proposal-level representation carrying fine-grained temporal information, $\sigma$ is the ReLU function, and  $\mathbf{W}_1^L,\mathbf{W}_2^L\in\mathbb{R}^{{(\textit{D}}/2)\times{\textit{D}}}$ are trainable weights. In practice, we find considering the features pooled from all snippets as a special snippet and including it in $S(i)$ marginally improves the performance.

\subsection{G-Net (Global Context)}
\label{sec:gnet}
The video-level global context is important as it involves background and high-level activity information that can be critical to distinguish action categories of similar appearance and motion patterns. A straightforward way to enrich the global context is to concatenate the video-level representation $\mathbf{z}$ and features of each proposal. However, this is insufficient because the global representation contains not only relevant context but also irrelevant noise. In addition, the contexts needed to process different proposals are different. This line of analysis motivates us to adapt the global context to each proposal before integrating it.

\textbf{Global context adaptation.}
To adapt the global context to the $i$th proposal, we first attend the video-level representation $\mathbf{z}$ to the features of this proposal $\mathbf{y}_i$ as well as the snippet-level features within it $\{\mathbf{x}_j: j\in S(i)\}$:

\begin{gather}
     a_{i,j}^G=\frac{\sigma(s(\mathbf{z},\mathbf{x}_j))}{\sum_{k\in{S(i)}} \sigma(s(\mathbf{z},\mathbf{x}_k))+\sigma(s(\mathbf{z},\mathbf{y}_i))},\\
     b_{i}^G=\frac{\sigma(s(\mathbf{z},\mathbf{y}_i))}{\sum_{k\in{S(i)}} \sigma(s(\mathbf{z},\mathbf{x}_k))+\sigma(s(\mathbf{z},\mathbf{y}_i))},\label{9}
\end{gather}
where $a_{i,j}^G$ is the attention weight between the video-level representation and the $j$th snippet of the $i$th proposal, $ b_{i}^G$ is the attention weight between the video-level representation and the $i$th proposal. 
With the attention weights, the adapted global context is calculated via
\begin{gather}
	\mathbf{z}_{i}^{G} = \sigma(\mathbf{W}_1^G\mathbf{z} + \mathbf{W}_2^G(\sum_{j \in S(i)} a_{i,j}^G\mathbf{x}_j + b_{i}^G\mathbf{y}_i)),\label{10}
\end{gather}
where $\mathbf{z}_{i}^{G}\in\mathbb{R}^{{\textit{D}}/2}$ is the global context adapted to the $i$th proposal, $\mathbf{W}_1^G,\mathbf{W}_2^G\in\mathbb{R}^{(\textit{D}/2)\times{\textit{D}}}$ are trainable weights.

\textbf{Global context aggregation.} 
Finally, G-Net concatenates the global context adapted to the $i$th proposal, \ie, $\mathbf{z}_{i}^{G}\in\mathbb{R}^{{\textit{D}}/2}$, and the features of this proposal obtained from L-Net, \ie, $\mathbf{y}_{i}^{L}\in\mathbb{R}^{{\textit{D}}/2}$ :
\begin{equation}
	\mathbf{y}_{i}^{G} = {\mathbf{y}_{i}^{L}}\oplus{\mathbf{z}_{i}^{G}},\label{1}
\end{equation}
where $\oplus$ denotes concatenation, and $\mathbf{y}_{i}^{G}\in\mathbb{R}^{{\textit{D}}}$ is the output of G-Net for the $i$th proposal.

\subsection{P-Net (Inter-proposal Relations)}
\label{sec:pnet}
P-Net takes as input $\{\mathbf{y}_{i}^{G}\}$ and outputs a new representation for each proposal.  
We use an existing model as P-Net and investigate two candidates: P-GCN~\cite{Alpher03} and the non-local network ~\cite{nonlocalnn}. 

P-GCN constructs an action proposal graph. Each proposal is
taken as a node. There are two types of relational edges. 
One connects overlapping proposals, and the other connects distinct but nearby proposals.
Then, a GCN is applied to update the proposal-level features based on their relations. Different from P-GCN, the non-local network builds a complete graph over all proposals and dynamically calculates the edge weights based on their pairwise similarities. We will have an in-depth investigation on their effectiveness as a building block in our ContextLoc.

Since each $\mathbf{y}_{i}^{G}$ is a concatenation of two parts as indicated in Eq. \eqref{1}, P-Net models not only interactions between proposal-level features enhanced  by  the  local  context  but also  interactions  between global contexts adapted to different proposals. We will show in the experiments the latter help improve the TAL performance.
Note we do not claim any specific network architecture we adopt in P-Net as our contribution. What we would like to show is that P-Net, regardless of its instantiation, is complementary to the local and global contexts we advocate in this paper and serves as a useful component in the proposed ContextLoc.

\begin{figure}[t]
\begin{center}
\includegraphics[width=1\linewidth]{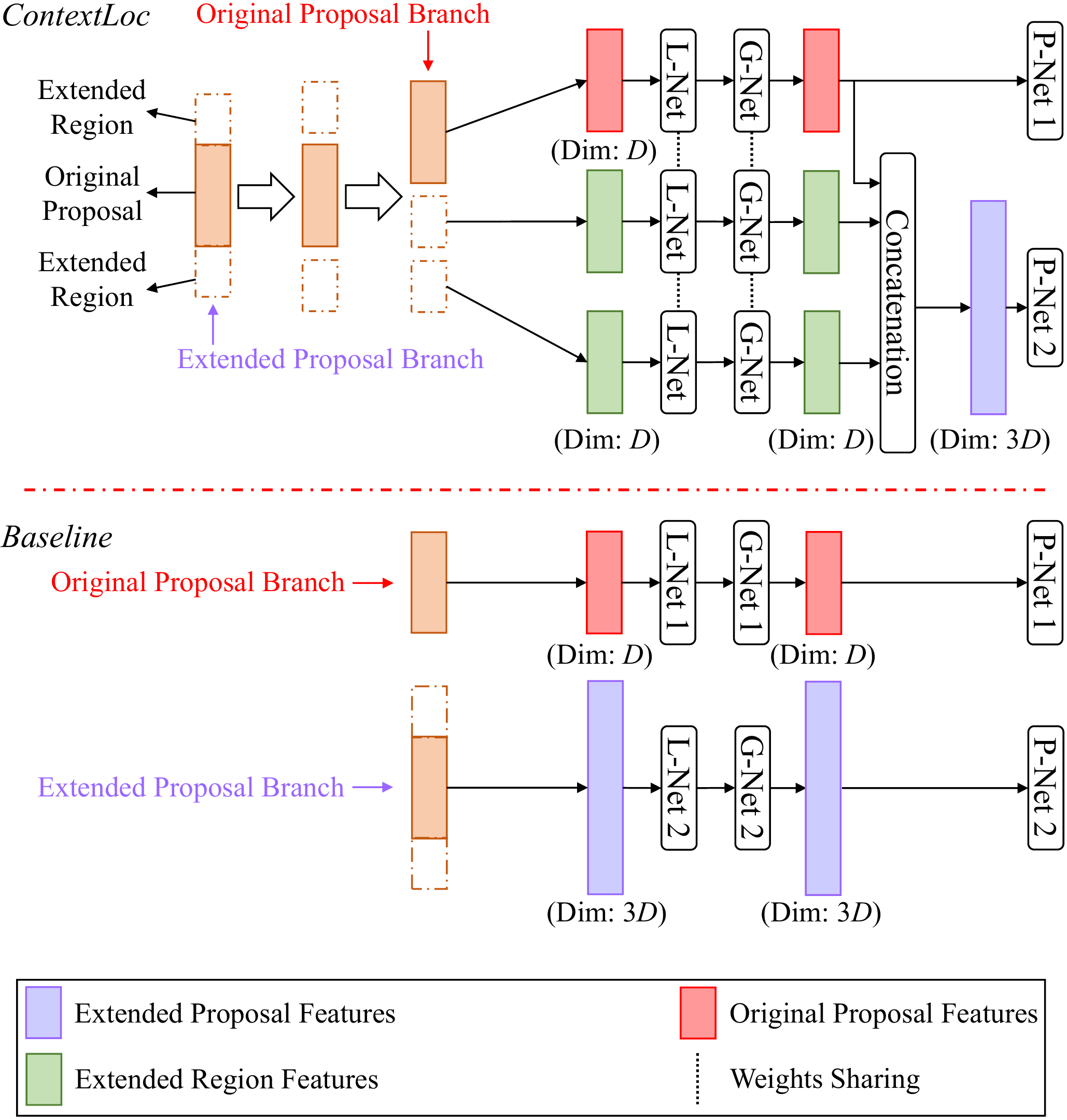}
\end{center}
   \caption{
Different methods of processing the extended proposals. To better highlight the original and extended proposals, we do not include other entities (\eg., snippets and the video) in this figure. Bottom: prior approaches treat an extended proposal as a single proposal and process it in a separate branch. Top: our new method treats the extended proposal as three proposals in L-Net and G-Net. P-Net, \ie, ``P-Net 2", treats an extended proposal as a single proposal and processes the concatenated features.
   }
\label{fig3}
\end{figure}
\subsection{Extended Proposals}
\label{sec:extend}
A common practice in TAL \cite{Shou2017CDCCN,Alpher03,Alpher02,Lin2020FastLO} is to extend each proposal on both ends (\eg, by 50\% of the temporal duration). The predictions obtained from these extended proposals and the original ones are fused as described in Section \ref{sec:overview}. Prior methods like P-GCN treat an extended proposal as a single proposal independent of the original proposal.
As we will show in the experiments, this strategy does not work well for L-Net and G-Net.
On the one hand, setting the feature dimension of an extended proposal much larger than that of the original proposal can account for the enlarged temporal duration but significantly increases the model complexity. On the other hand, setting the feature dimension of an extended proposal the same as that of the original proposal leads to inferior performance.
In addition, the extended proposal and the original one are processed separately. Their internal connection (\ie, the original proposal is a part of the extended proposal) is ignored.

To address this problem, we treat the extended proposal as three proposals in L-Net and G-Net, \ie, the original proposal and the extended regions on the two sides. The duration of each extended region is 50\% as long as that of the original proposal. As shown in Figure \ref{fig3}, L-Net and G-Net process  these three proposals separately but use shared weights. Then, we concatenate their new representations. Finally, P-Net, \ie, ``P-Net 2'' in Figure \ref{fig3} Top, treats an extended proposal as a single proposal and processes these concatenated features. Note the processing of the original proposal in L-Net and G-Net is now a part of the processing of the extended proposal. This not only reduces the model size and computational complexity but also reflects the connection between the original proposal and the extended one.

\section{Experiments}\label{ex}



\textbf{Datasets.} The THUMOS14~\cite{t14} dataset contains a large number of open source videos of human actions in real environments. It includes four parts: the training set, verification set, background set, and test set. The training set is based on the UCF-101 dataset~\cite{Soomro2012UCF101AD}, and the videos are edited (each video usually contains an instance). For the remaining three parts (verification, background and test sets), the videos are untrimmed. Following the common setting in THUMOS14, we apply 200 videos in the validation set for training and conduct evaluation on the 213 annotated videos from the test set. ActivityNet v1.3~\cite{Heilbron2015ActivityNetAL} is currently the largest dataset of action analysis in videos, including 20,000 Youtube videos with 200 action categories. The training set contains about 10,000 videos. Both the validation set and the test set contain about 5,000 videos. On average, each video has 1.5 action instances. Following the standard practice, we train our method on the training videos and test it on the validation videos.


\textbf{Evaluation metric.} We evaluate the performance of ContextLoc  using mean average precision (mAP) values at different tIoU thresholds. On THUMOS14, the tIoU thresholds are chosen
from \{0.3, 0.4, 0.5, 0.6, 0.7\}. On ActivityNet v1.3, the tIoU
thresholds are from \{0.5, 0.75, 0.95\}, and we also report the
average mAP of the tIoU thresholds between 0.5 and 0.95 with a step of 0.05.

\textbf{Implementation details.} We divide each input video into 64-frame snippets and obtain snippet-level features (1024-dimensional feature vectors) via the I3D network~\cite{Alpher01}. 
We implement ContextLoc using PyTorch. It is trained on three NVIDIA GeForce GTX TITAN XP GPUs with a batch size of 32. The stochastic gradient descent (SGD) solver is adopted for optimization. On THUMOS14, the initial learning rate is 0.01 and it is divided by 10 after 15 epochs. On ActivityNet v1.3, the initial learning rate is 0.01 for the first 15 epochs and it is divided by 10 after 10 epochs. The final score for calculating mAP is obtained by multiplying the scores of action classification and completeness prediction. The predictions from the RGB and optical flow steams are fused with the ratio of 5:6. Non-Maximum Suppression (NMS) removes duplicated action proposals and obtains the final predicted proposals. Unless otherwise stated, P-GCN is taken as the P-Net.

\textbf{Loss functions.} The action classification of original proposals is trained with the cross-entropy loss, the completeness prediction is trained with the hinge loss, and the boundary regression is trained with the smooth $L_1$ loss. 

\begin{table}[t]
\begin{center}
\scalebox{0.90}{
\begin{tabular}{l|c|c|c|c}
\hline
Method & RGB & Flow & Fusion & tIoU@0.5\\
\hline\hline
P-Net (non-local~\cite{nonlocalnn}) & 32.16 & 35.13 & 39.76 & 45.81 \\
+L-Net & 34.48 & 38.13 & 42.64 & 50.46 \\
+G-Net & 34.31 & 36.97 & 41.33 & 47.74 \\
+L-Net + G-Net & 35.77 & 40.07 & 44.03 & 51.00\\
\hline
P-Net (P-GCN~\cite{Alpher03}) & 34.93 & 39.26 & 42.43 & 49.10 \\
+L-Net & 36.40 & 40.48 & 45.12 & 53.17 \\
+G-Net & 36.15 & 40.42 & 43.51 & 50.11 \\
+L-Net + G-Net & 37.23 & 42.52 & 45.70 & 54.30\\
\hline
\end{tabular}}
\end{center}
\caption{Ablation study on the effectiveness of L-Net and G-Net. P-Net is taken as a strong baseline. The average mAP ($\%$) of tIoU thresholds from 0.1 to 0.9 and the mAP of tIoU@0.5 are reported on the THUMOS14 test set.}
\label{SPGCCNex}
\end{table}


\subsection{Ablation Study}
To explore the effectiveness of each component under a controlled setting, we remove or change the corresponding part of ContextLoc  in each ablation experiment while keeping the others (not mentioned) the same. All results are reported on THUMOS14.
\begin{table}[t]
\begin{center}
\scalebox{1}{
\begin{tabular}{c c|c|c|c|c}
\hline
Adapt. & Inter. & RGB & Flow & Fusion & Class\%\\
\hline\hline
  &  & 34.98 & 39.45 & 42.27 & 76.2\\
  $\checkmark$ &  & 37.15 & 40.62 & 44.14 & 77.8\\
  & $\checkmark$ & 35.26 & 41.05 & 43.23 & 76.8\\
 $\checkmark$ & $\checkmark$ & 37.23 & 42.52 & 45.70 & 78.5\\
\hline
\end{tabular}}
\end{center}
\caption{
Ablation study on the individual and combinatorial effects of global context adaptation in G-Net (denoted as ``Adapt.'') and interactions between adapted global contexts achieved by P-Net (denoted as ``Inter.''). 
The average mAP ($\%$) of tIoU thresholds from 0.1 to 0.9 and the action classification accuracy (Class\%) are reported on the THUMOS14 test set.}
\label{times}
\end{table}

\textbf{ContextLoc sub-networks.} 
Since prior work has already proved that modeling inter-proposal relations helps TAL, we take P-Net as a strong baseline to validate the effectiveness of L-Net (local context) and G-Net (global context), which are the main contribution of this paper. We consider two instantiations of P-Net, \ie, P-GCN and the non-local network. The results are shown in Table \ref{SPGCCNex}.

When the non-local network is taken as P-Net, adding L-Net alone before it improves 2.88\% on average mAP and 4.65\% on tIoU 0.5. 
Adding G-Net alone before P-Net improves 1.57\% on average mAP and 1.93\% on tIoU 0.5. 
Our complete ContextLoc (L-Net+G-Net+P-Net)  further improves the performance.

When P-GCN is taken as P-Net, adding L-Net alone before it improves $2.69\%$ on average mAP and $4.07\%$ on tIoU 0.5. 
Adding G-Net alone before P-Net improves $1.08\%$ on average mAP and $1.01\%$ on tIoU 0.5. 
The complete ContextLoc (L-Net+G-Net+P-Net)  further improves the performance. 

Comparing the two sets of results achieved via different instantiations of P-Net, we can observe that P-GCN outperforms the non-local network regardless of whether P-Net works alone or is integrated with other sub-networks. 
The reason is that the non-local network is not specially designed for TAL.
Therefore, in all other experiments, we will use P-GCN as the P-Net.

Several conclusions can be drawn from this experiment. (1) Either the local context modeled via L-Net or the global context modeled via G-Net benefits TAL.  (2) The local context, global context and inter-proposal relations are complementary. Our ContextLoc unifies their advantages and achieves the best performance. (3) The effectiveness of L-Net, G-Net and ContextLoc does not depend on any specific instantiation of P-Net. They consistently improve the performance over the strong baselines.

\begin{table}[t]
\begin{center}
\scalebox{0.93}{
\begin{tabular}{l |c|c|c|c|c}
\hline
 Method & Dim. & Fusion & tIoU@0.5 &\#Params &Flops \\
\hline\hline
Baseline & 1024 & 40.04 & 45.87 & 6.5M & 3.6G \\
Baseline & 3072 & 44.57 & 52.68 & 25.6M & 16.2G\\
Ours & 3072 & 45.70 & 54.30 & 6.7M & 3.1G\\
\hline
\end{tabular}}
\end{center}
\caption{
Ablation study on different methods of processing extended proposals (as illustrated in Figure \ref{fig3}). ``Dim.'' denotes the dimension of features of an extended proposal.
The average mAP ($\%$) of tIoU thresholds from 0.1 to 0.9 and the mAP of tIoU@0.5 are reported on the THUMOS14 test set.}
\label{extp}
\end{table}

\begin{table}[t]
\begin{center}
\scalebox{0.85}{
\begin{tabular}{l |c|c|c|c|c}
\hline
Method & Dim. & Fusion & tIoU@0.5 & \#Params & Flops\\
\hline\hline
P-GCN & 1024 & 42.43 & 49.10 & 4.6M & 1.7G \\
Deep P-GCN & 1024 & 42.20 & 48.76 & 6.4M & 2.6G \\
Deeper P-GCN & 1024 & 42.03 & 48.86 & 8.8M & 3.3G \\
\hline
ContextLoc & 1024 & 45.70 & 54.30 & 6.7M & 3.1G  \\
ContextLoc & 512  & 44.27 & 52.32 & 3.4M & 1.6G \\
\hline
\end{tabular}}
\end{center}
\caption{Comparison of model sizes and complexities. Deep P-GCN and Deeper P-GCN respectively add one and two graph convolutional layers to P-GCN. ``Dim." denotes the dimension of features of a proposal, \ie, $D$. The average mAP ($\%$) is calculated on tIoU thresholds from 0.1 to 0.9.  
Results are reported on the THUMOS14 test set. }
\label{modelsize}
\end{table}

\begin{table}[t]
	\begin{center}
\scalebox{1}{
	\begin{tabular}{l |c|c}
		\hline
		Method & Class\% & tIoU\% \\
		\hline\hline
		P-Net (P-GCN) & 74.3 & 78.2 \\
		+ L-Net & 75.9 & 81.4  \\
		+ G-Net & 77.4 & 79.1 \\
		+ L-Net + G-Net & 78.5 & 82.0 \\
		\hline
\end{tabular}}

	\end{center}
\caption{Ablation study on classification and localization. The action classification accuracy (Class\%) and tIoU w.r.t. ground truth (tIoU\%) after regression are reported on the THUMOS14 test set.}
\label{tab_twotables}
\end{table}



\textbf{Global context adaptation \& interactions between adapted global contexts.} 
This ablation study investigates the individual and combinatorial effects of global context adaption in G-Net and interactions between adapted global contexts achieved by P-Net.
To deactivate the latter, we let global context aggregation occur after the proposal-level features pass through P-Net.
Specifically, we first send proposal-level features obtained from L-Net, \ie, $\{\mathbf{y}_i^L\}$, to P-Net and then concatenate the new representations of each proposal with their corresponding adapted global contexts, \ie, $\{\mathbf{z}_i^G\}$. 

The results are shown in Table \ref{times}.
We can see that both global context adaptation and interactions between adapted global contexts improve the performance. The former benefits the RGB stream more than the flow stream while the latter play a more important role in the flow stream. They together help ContextLoc achieve the best performance.

\textbf{Extended proposals.}
Prior approaches process the extended proposal as a single proposal. 
We instead process it as three proposals in L-Net and G-Net and process the concatenation of updated features in P-Net. Results are shown in Table \ref{extp}.
When the feature dimension of an extended proposal in the baseline method is the same as that of the original proposal, \ie, 1024, its performance is much worse than that of our new method.
When the feature dimension of an extended proposal in the baseline method is much larger than that of the original proposal, \ie, 3072, its performance gets better (with significantly more parameters and flops) but is still inferior to ours.



\textbf{Model size \& complexity.} 
The model size and computational complexity of our ContextLoc are 6.7M and 3.1G flops, respectively. In contract, P-GCN has 4.6M parameters and 1.7G flops. This is expected because ContextLoc includes P-Net as a building component.
To have a fair comparison, we construct a Deep P-GCN and a Deeper P-GCN by adding one and two graph convolutional layers to P-GCN, respectively. Table \ref{modelsize} shows that naively increasing the model complexity and network depth can not improve the performance. This further proves that the local and global contexts we advocate is complementary to the inter-proposal relations.


We also experiment on a light version of ContextLoc by halving the feature dimensions. With fewer parameters (3.4M versus 4.6M) and flops (1.6G versus 1.7G) than P-GCN, it outperforms P-GCN by $3.22\%$ at tIoU@0.5. This again verifies that the effectiveness of our ContextLoc is not caused by a larger model size or more flops.

\textbf{Concrete analysis.} Table \ref{tab_twotables} reports the classification and localization accuracy of high-quality proposals, whose tIoU w.r.t. ground truth are greater than 0.7. The results indicate L-Net contributes more to localization (tIoU\%) and G-Net more to classification, which function as expected.


\subsection{Comparison with State-of-the-Art Methods}

\begin{table}[t]
\begin{center}
\begin{tabular}{l|c c c c c}
\hline
Method & 0.3 & 0.4 & 0.5 & 0.6 & 0.7\\
\hline\hline
CDC~\cite{Shou2017CDCCN} & 40.1 & 29.4 & 23.3 & 13.1 & 7.9 \\
TCN~\cite{Dai2017TemporalCN} & - & 33.3 & 25.6 & 15.9 & 9.0 \\
TURN-TAP~\cite{Gao2017TURNTT} & 44.1 & 34.9 & 25.6 & - & - \\
R-C3D~\cite{Xu2017RC3DRC} & 44.8 & 35.6 & 28.9 & - & - \\
SS-TAD~\cite{Buch2017EndtoEndST} & 45.7 & - & 29.2 & - & 9.6 \\
SSN~\cite{zhao2017temporal} & 51.9 & 41.0 & 29.8 & - & - \\
BSN~\cite{Alpher02} & 53.5 & 45.0 & 36.9 & 28.4 & 20.0 \\
MGG~\cite{onestage2} & 53.9 & 46.8 & 37.4 & 29.5 & 21.3 \\
BMN~\cite{bmn} & 56.0 & 47.4 & 38.8 & 29.7 & 20.5 \\
G-TAD~\cite{gtad} & 54.5 & 47.6 & 40.2 & 30.8 & 23.4 \\
TAL-Net~\cite{Chao2018RethinkingTF} & 53.2 & 48.5 & 42.8 & 33.8 & 20.8\\
Huang \etal~\cite{Huang2019DecouplingLA} & 60.2 & 54.1 & 44.2 & 32.3 & 19.1\\
Zhao \etal~\cite{zhao2020bottom} & 53.9 & 50.7 & 45.4 & 38.0 & 28.5\\
P-GCN~\cite{Alpher03} & 63.6 & 57.8 & 49.1 & - & -\\
PBRNet~\cite{Liu2020ProgressiveBR} & 58.5 & 54.6 & 51.3 & \textbf{41.8} & \textbf{29.5} \\
ContextLoc (ours) & \textbf{68.3} & \textbf{63.8} & \textbf{54.3} & \textbf{41.8} & 26.2\\
\hline
\end{tabular}
\end{center}
\caption{Results on the THUMOS14 test set. The mAP ($\%$) at different tIoU thresholds are reported. \textbf{Bold} fonts indicate the best performance.}
\label{t14ex}
\end{table}

\begin{table}[t]
\begin{center}
\begin{tabular}{l|c c c c}
\hline
Method & 0.5 & 0.75 & 0.95 & Average\\
\hline\hline
R-C3D~\cite{Xu2017RC3DRC} & 26.80 & - & - & -\\
TAL-Net~\cite{Chao2018RethinkingTF} & 38.23 & 18.30 & 1.30 & 20.22 \\
SSN~\cite{zhao2017temporal} & 43.26 & 28.70 & 5.63 & 28.28 \\
BSN~\cite{Alpher02} & 46.45 & 29.96 & 8.02 & 30.03 \\
BMN~\cite{bmn} & 50.07 & 34.78 & 8.29 & 33.85\\
G-TAD~\cite{gtad} & 50.36 & 34.60 & 9.02 & 34.09  \\
Zhao \etal~\cite{zhao2020bottom} & 43.47 & 33.91 &\textbf{9.21} & 30.12 \\
PBRNet~\cite{Liu2020ProgressiveBR} & 53.96 & 34.97 & 8.98 & \textbf{35.01}\\
P-GCN~\cite{Alpher03} & 42.90 & 28.14 & 2.47 & 26.99 \\
ContextLoc (ours) & 51.24 & 31.40 & 2.83 & 30.59\\
\hline
P-GCN*~\cite{Alpher03} & 48.26 & 33.16 & 3.27 & 31.11 \\
ContextLoc* (ours) & \textbf{56.01} & \textbf{35.19} & 3.55 & 34.23\\
\hline
\end{tabular}
\end{center}
\caption{Results on the ActivityNet v1.3 validation set. The mAP ($\%$) at different tIoU thresholds and the average mAP of IoU thresholds from 0.5 to 0.95 are reported. Symbol (*) denotes using external video-level classes predictions. \textbf{Bold} fonts indicate the best performance.}
\label{anet13ex}
\end{table}

\textbf{THUMOS14.} As summarized in Table \ref{t14ex}, our ContextLoc outperforms all the compared methods on the THUMOS14 test set. At tIoU 0.5, ContextLoc outperforms the previously best method PBRNet~\cite{Liu2020ProgressiveBR} by $3.0\%$ absolute improvement. 
This demonstrates the benefit of enriching local and global contexts for TAL.

\textbf{ActivityNet v1.3.} 
The results on ActivityNet v1.3 are shown in Table \ref{anet13ex}.
Zeng \etal~\cite{Alpher03} found taking into account video-level classes predicted by UntrimmedNet \cite{twostage1} helps improve the performance of P-GCN. We also adopt the same strategy to obtain ContextLoc*. Specifically, we add a branch of video-level representation $\mathbf{z}$ to predict the video-level classes. More details about ContextLoc* can be found in the supplementary material. At tIoU 0.5, ContextLoc* reaches an mAP of $56.01\%$ which is $2.05\%$ higher than the current best $53.96\%$ achieved by PBRNet~\cite{Liu2020ProgressiveBR}. At tIoU 0.75, ContextLoc* outperforms the previously best method PBRNet by $0.22\%$. At tIoU 0.95, ContextLoc* does not perform as well as previous methods and only marginally outperforms P-GCN. In the supplementary material, we visualized some results to explain this gap.

\begin{figure}[t]
\centering
  \subfigure[Qualitative results on an example from the THUMOS14 test set (top) and an example from the ActivityNet v1.3 validation set (bottom) show that our ContextLoc locates the temporal boundaries more accurately than P-GCN. ]{\includegraphics[width=1\linewidth]{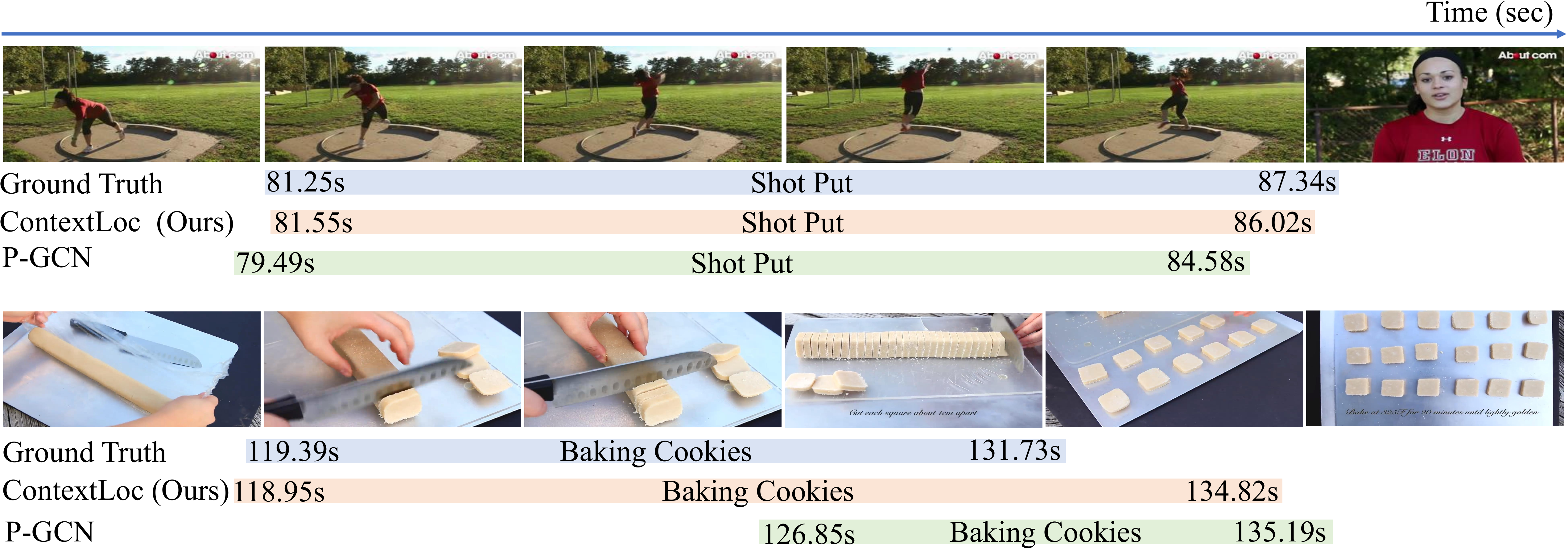}
  \label{fig4}}
  \subfigure[Qualitative results on an example from the THUMOS14 test set (top) and an example from the ActivityNet v1.3 validation set (bottom) show that our ContextLoc correctly classifies action instances which P-GCN fails on.]{\includegraphics[width=1\linewidth]{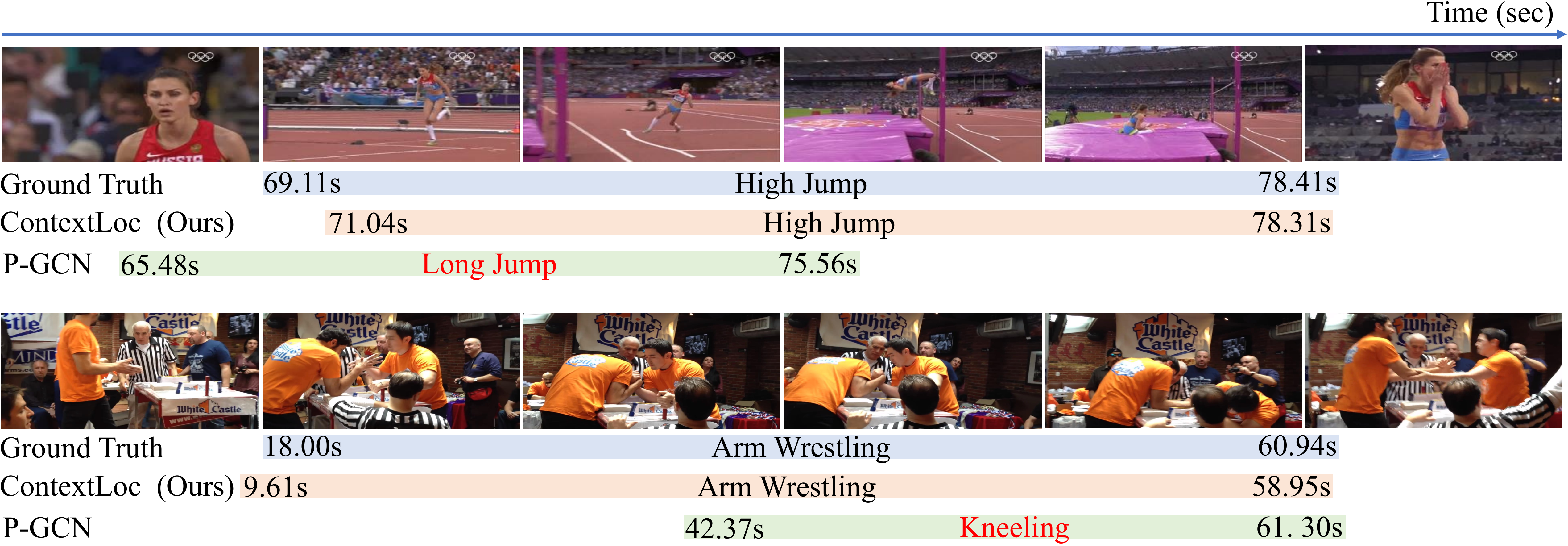}
  \label{fig10}}
\caption{
Results from the ground truth, ContextLoc and P-GCN are respectively illustrated using blue, orange and green bars.}
\end{figure}

\subsection{Qualitative Results}

To validate the effectiveness of temporal boundary location and action classification, we visualize qualitative results in Figure \ref{fig4} and Figure \ref{fig10}, respectively. Compared with P-GCN, ContextLoc predicts a more precise temporal boundary and classification for these examples.

\section{Conclusion}
This paper introduces a novel network architecture, termed \textit{ContextLoc}, for TAL. It models the local context, global context and context-aware inter-proposal relations in a unified framework. 
Ablation experiments under controlled settings indicate the effectiveness of each component of ContextLoc. Results on two datasets demonstrate that ContextLoc outperforms state-of-the-art TAL methods.

\textbf{Acknowledgment.} This work was supported partly by National Key R\&D Program of China Grant 2018AAA0101400, NSFC Grants 62088102, 61976171, and XXXXXXXX, and Young Elite Scientists Sponsorship Program by CAST Grant 2018QNRC001.

{\small
\bibliographystyle{ieee_fullname}
\bibliography{egbib}

\begin{thebibliography}{10}\itemsep=-1pt

\bibitem{bai2020boundary}
Yueran Bai, Yingying Wang, Yunhai Tong, Yang Yang, Qiyue Liu, and Junhui Liu.
\newblock Boundary content graph neural network for temporal action proposal
  generation.
\newblock In {\em ECCV}, pages 121--137, 2020.

\bibitem{Buch2017EndtoEndST}
Shyamal Buch, Victor Escorcia, Bernard Ghanem, Li Fei-Fei, and Juan~Carlos
  Niebles.
\newblock End-to-end, single-stream temporal action detection in untrimmed
  videos.
\newblock In {\em BMVC}, 2017.

\bibitem{Alpher01}
Joao Carreira and Andrew Zisserman.
\newblock Quo vadis, action recognition? a new model and the kinetics dataset.
\newblock In {\em CVPR}, pages 6299--6308, 2017.

\bibitem{Chao2018RethinkingTF}
Yu-Wei Chao, Sudheendra Vijayanarasimhan, Bryan Seybold, David~A. Ross, Jia
  Deng, and Rahul Sukthankar.
\newblock Rethinking the faster r-cnn architecture for temporal action
  localization.
\newblock In {\em CVPR}, pages 1130--1139, 2018.

\bibitem{Chen2020RefinementOB}
Yunze Chen, Mengjuan Chen, Rui Wu, Jiagang Zhu, Zheng Zhu, and Qingyi Gu.
\newblock Refinement of boundary regression using uncertainty in temporal
  action localization.
\newblock In {\em BMVC}, 2020.

\bibitem{chen2020hierarchical}
Zhao-Min Chen, Xin Jin, Borui Zhao, Xiu-Shen Wei, and Yanwen Guo.
\newblock Hierarchical context embedding for region-based object detection.
\newblock In {\em ECCV}, pages 633--648, 2020.

\bibitem{Dai2017TemporalCN}
Xiyang Dai, Bharat Singh, Guyue Zhang, Larry~S. Davis, and Yan~Qiu Chen.
\newblock Temporal context network for activity localization in videos.
\newblock In {\em ICCV}, pages 5727--5736, 2017.

\bibitem{fan2019heterogeneous}
Chenyou Fan, Xiaofan Zhang, Shu Zhang, Wensheng Wang, Chi Zhang, and Heng
  Huang.
\newblock Heterogeneous memory enhanced multimodal attention model for video
  question answering.
\newblock In {\em CVPR}, pages 1999--2007, 2019.

\bibitem{Gallese1996ActionRI}
Vittorio Gallese, Luciano Fadiga, Leonardo Fogassi, and Giacomo Rizzolatti.
\newblock Action recognition in the premotor cortex.
\newblock {\em Brain : a journal of neurology}, 119 (Pt 2):593--609, 1996.

\bibitem{onesstage1}
Jiyang Gao, Kan Chen, and Ram Nevatia.
\newblock Ctap: Complementary temporal action proposal generation.
\newblock In {\em ECCV}, pages 68--83, 2018.

\bibitem{Gao2017TURNTT}
Jiyang Gao, Zhenheng Yang, Chen Sun, Kan Chen, and Ram Nevatiawqer.
\newblock Turn tap: Temporal unit regression network for temporal action
  proposals.
\newblock In {\em ICCV}, pages 3648--3656, 2017.

\bibitem{he2019stnet}
Dongliang He, Zhichao Zhou, Chuang Gan, Fu Li, Xiao Liu, Yandong Li, Limin
  Wang, and Shilei Wen.
\newblock Stnet: Local and global spatial-temporal modeling for action
  recognition.
\newblock In {\em AAAI}, pages 8401--8408, 2019.

\bibitem{Heilbron2015ActivityNetAL}
Fabian~Caba Heilbron, Victor Escorcia, Bernard Ghanem, and Juan~Carlos Niebles.
\newblock Activitynet: A large-scale video benchmark for human activity
  understanding.
\newblock In {\em CVPR}, pages 961--970, 2015.

\bibitem{Huang2019DecouplingLA}
Yupan Huang, Qi Dai, and Yutong Lu.
\newblock Decoupling localization and classification in single shot temporal
  action detection.
\newblock In {\em ICME}, pages 1288--1293, 2019.

\bibitem{Ji20133DCN}
Shuiwang Ji, Wei Xu, Ming Yang, and Kai Yu.
\newblock 3d convolutional neural networks for human action recognition.
\newblock {\em IEEE Transactions on Pattern Analysis and Machine Intelligence},
  35:221--231, 2013.

\bibitem{t14}
Yu-Gang Jiang, Haroon Idrees, Amir Zamir, Alex Gorban, Ivan Laptev, Mubarak
  Shah, and Rahul Sukthankar.
\newblock Thumos challenge: Action recognition with a large number of classes.
\newblock 2014.

\bibitem{kay2017kinetics}
Will Kay, Joao Carreira, Karen Simonyan, Brian Zhang, Chloe Hillier, Sudheendra
  Vijayanarasimhan, Fabio Viola, Tim Green, Trevor Back, Paul Natsev, et~al.
\newblock The kinetics human action video dataset.
\newblock {\em ArXiv}, abs/1705.06950, 2017.

\bibitem{otherareas4}
Dahun Kim, Donghyeon Cho, and In~So Kweon.
\newblock Self-supervised video representation learning with space-time cubic
  puzzles.
\newblock In {\em AAAI}, pages 8545--8552, 2019.

\bibitem{Lin2020FastLO}
Chuming Lin, Jian Li, Yabiao Wang, Ying Tai, Donghao Luo, Zhipeng Cui, Chengjie
  Wang, Jilin Li, Feiyue Huang, and Rongrong Ji.
\newblock Fast learning of temporal action proposal via dense boundary
  generator.
\newblock In {\em AAAI}, 2020.

\bibitem{bmn}
Tianwei Lin, Xiao Liu, Xin Li, Errui Ding, and Shilei Wen.
\newblock Bmn: Boundary-matching network for temporal action proposal
  generation.
\newblock In {\em ICCV}, pages 3888--3897, 2019.

\bibitem{Alpher02}
Tianwei Lin, Xu Zhao, Haisheng Su, Chongjing Wang, and Ming Yang.
\newblock Bsn: Boundary sensitive network for temporal action proposal
  generation.
\newblock In {\em ECCV}, pages 3--19, 2018.

\bibitem{Liu_2020_CVPR}
Chenchen Liu, Yang Jin, Kehan Xu, Guoqiang Gong, and Yadong Mu.
\newblock Beyond short-term snippet: Video relation detection with
  spatio-temporal global context.
\newblock In {\em CVPR}, pages 10837--10846, 2020.

\bibitem{Liu2020ProgressiveBR}
Qinying Liu and Zilei Wang.
\newblock Progressive boundary refinement network for temporal action
  detection.
\newblock In {\em AAAI}, 2020.

\bibitem{liu2018global}
Yang Liu, Zhaoyang Lu, Jing Li, Tao Yang, and Chao Yao.
\newblock Global temporal representation based cnns for infrared action
  recognition.
\newblock {\em IEEE Signal Processing Letters}, 25(6):848--852, 2018.

\bibitem{onestage2}
Yuan Liu, Lin Ma, Yifeng Zhang, Wei Liu, and Shih-Fu Chang.
\newblock Multi-granularity generator for temporal action proposal.
\newblock In {\em CVPR}, pages 3604--3613, 2019.

\bibitem{long2019gaussian}
Fuchen Long, Ting Yao, Zhaofan Qiu, Xinmei Tian, Jiebo Luo, and Tao Mei.
\newblock Gaussian temporal awareness networks for action localization.
\newblock In {\em CVPR}, pages 344--353, 2019.

\bibitem{otherareas3}
Thien Nguyen and Ralph Grishman.
\newblock Graph convolutional networks with argument-aware pooling for event
  detection.
\newblock In {\em AAAI}, pages 5900--5907, 2018.

\bibitem{qiu2019learning}
Zhaofan Qiu, Ting Yao, Chong-Wah Ngo, Xinmei Tian, and Tao Mei.
\newblock Learning spatio-temporal representation with local and global
  diffusion.
\newblock In {\em CVPR}, pages 12056--12065, 2019.

\bibitem{ren2016faster}
Shaoqing Ren, Kaiming He, Ross Girshick, and Jian Sun.
\newblock Faster r-cnn: towards real-time object detection with region proposal
  networks.
\newblock {\em IEEE Transactions on Pattern Analysis and Machine Intelligence},
  39(6):1137--1149, 2016.

\bibitem{Shou2017CDCCN}
Zheng Shou, Jonathan Chan, Alireza Zareian, Kazuyuki Miyazawa, and Shih-Fu
  Chang.
\newblock Cdc: Convolutional-de-convolutional networks for precise temporal
  action localization in untrimmed videos.
\newblock In {\em CVPR}, pages 1417--1426, 2017.

\bibitem{Shou2016TemporalAL}
Zheng Shou, Dongang Wang, and Shih-Fu Chang.
\newblock Temporal action localization in untrimmed videos via multi-stage
  cnns.
\newblock In {\em CVPR}, pages 1049--1058, 2016.

\bibitem{Simonyan2014TwoStreamCN}
Karen Simonyan and Andrew Zisserman.
\newblock Two-stream convolutional networks for action recognition in videos.
\newblock In {\em NeurIPS}, pages 568--576, 2014.

\bibitem{Soomro2012UCF101AD}
Khurram Soomro, Amir~Roshan Zamir, and Mubarak Shah.
\newblock Ucf101: A dataset of 101 human actions classes from videos in the
  wild.
\newblock {\em ArXiv}, abs/1212.0402, 2012.

\bibitem{Tran2015LearningSF}
Du Tran, Lubomir Bourdev, Rob Fergus, Lorenzo Torresani, and Manohar Paluri.
\newblock Learning spatiotemporal features with 3d convolutional networks.
\newblock In {\em ICCV}, pages 4489--4497, 2015.

\bibitem{Tran2018ACL}
Du Tran, Heng Wang, Lorenzo Torresani, Jamie Ray, Yann LeCun, and Manohar
  Paluri.
\newblock A closer look at spatiotemporal convolutions for action recognition.
\newblock In {\em CVPR}, pages 6450--6459, 2018.

\bibitem{Vaswani2017AttentionIA}
Ashish Vaswani, Noam Shazeer, Niki Parmar, Jakob Uszkoreit, Llion Jones,
  Aidan~N. Gomez, Łukasz Kaiser, and Illia Polosukhin.
\newblock Attention is all you need.
\newblock In {\em NeurIPS}, pages 5999--6009, 2017.

\bibitem{otherareas2}
Heng Wang, Dan Oneata, Jakob Verbeek, and Cordelia Schmid.
\newblock A robust and efficient video representation for action recognition.
\newblock {\em International Journal of Computer Vision}, 119(3):219--238,
  2016.

\bibitem{Wang2013ActionRW}
Heng Wang and Cordelia Schmid.
\newblock Action recognition with improved trajectories.
\newblock In {\em ICCV}, pages 3551--3558, 2013.

\bibitem{twostage1}
Limin Wang, Yuanjun Xiong, Dahua Lin, and Luc Van~Gool.
\newblock Untrimmednets for weakly supervised action recognition and detection.
\newblock In {\em CVPR}, pages 4325--4334, 2017.

\bibitem{nonlocalnn}
Xiaolong Wang, Ross Girshick, Abhinav Gupta, and Kaiming He.
\newblock Non-local neural networks.
\newblock In {\em CVPR}, pages 7794--7803, 2018.

\bibitem{Xu2017RC3DRC}
Huijuan Xu, Abir Das, and Kate Saenko.
\newblock R-c3d: Region convolutional 3d network for temporal activity
  detection.
\newblock In {\em ICCV}, pages 5794--5803, 2017.

\bibitem{gtad}
Mengmeng Xu, Chen Zhao, David~S. Rojas, Ali Thabet, and Bernard Ghanem.
\newblock G-tad: Sub-graph localization for temporal action detection.
\newblock In {\em CVPR}, pages 10153--10162, 2020.

\bibitem{yeung2016end}
Serena Yeung, Olga Russakovsky, Greg Mori, and Li Fei-Fei.
\newblock End-to-end learning of action detection from frame glimpses in
  videos.
\newblock In {\em CVPR}, pages 2678--2687, 2016.

\bibitem{Alpher03}
Runhao Zeng, Wenbing Huang, Mingkui Tan, Yu Rong, Peilin Zhao, Junzhou Huang,
  and Chuang Gan.
\newblock Graph convolutional networks for temporal action localization.
\newblock In {\em CVPR}, pages 7094--7103, 2019.

\bibitem{zhao2020bottom}
Peisen Zhao, Lingxi Xie, Chen Ju, Ya Zhang, Yanfeng Wang, and Qi Tian.
\newblock Bottom-up temporal action localization with mutual regularization.
\newblock In {\em ECCV}, pages 539--555, 2020.

\bibitem{zhao2017temporal}
Yue Zhao, Yuanjun Xiong, Limin Wang, Zhirong Wu, Xiaoou Tang, and Dahua Lin.
\newblock Temporal action detection with structured segment networks.
\newblock In {\em ICCV}, pages 2914--2923, 2017.

\end{thebibliography}
}

\end{document}